\documentclass[final,journal]{IEEEtran}
\usepackage{cite}

\ifCLASSINFOpdf
   \usepackage[pdftex]{graphicx}
\else
   \usepackage[dvips]{graphicx}
\fi
\usepackage[cmex10]{amsmath}
\usepackage{algorithm}
\usepackage[noend]{algorithmic}
\renewcommand{\algorithmicrequire}{\textbf{Input:}}

\usepackage{array}
\usepackage{fixltx2e}
\usepackage{url}

\usepackage{enumitem}
\setlist{leftmargin=*}
\renewcommand{\descriptionlabel}[1]{\hspace{\labelsep}\textsc{#1}}

\usepackage{color}

\begin{document}
%
\title{Particle Swarm Optimization for Time Series\\ Motif Discovery}
%
%
%

\author{Joan~Serr\`a and Josep~Lluis~Arcos
\thanks{J.~Serr\`a and J.~Ll.~Arcos are with IIIA-CSIC, the Artificial Intelligence Research Institute of the Spanish National Research Council, Campus de la UAB s/n, 08193 Bellaterra, Barcelona, Spain, email: \{jserra,arcos\}@iiia.csic.es.}
}

%



\maketitle

\begin{abstract}
Efficiently finding similar segments or motifs in time series data is a fundamental task that, due to the ubiquity of these data, is present in a wide range of domains and situations. Because of this, countless solutions have been devised but, to date, none of them seems to be fully satisfactory and flexible. In this article, we propose an innovative standpoint and present a solution coming from it: an anytime multimodal optimization algorithm for time series motif discovery based on particle swarms. By considering data from a variety of domains, we show that this solution is extremely competitive when compared to the state-of-the-art, obtaining comparable motifs in considerably less time using minimal memory. In addition, we show that it is robust to different implementation choices and see that it offers an unprecedented degree of flexibility with regard to the task. All these qualities make the presented solution stand out as one of the most prominent candidates for motif discovery in long time series streams. Besides, we believe the proposed standpoint can be exploited in further time series analysis and mining tasks, widening the scope of research and potentially yielding novel effective solutions.
\end{abstract}


%
\IEEEpeerreviewmaketitle

%
%
%
%

\section{Introduction}
\label{sec:intro}

\IEEEPARstart{T}{ime series} are sequences of real numbers measured at successive, usually regular time intervals. Data in the form of time series pervade science, business, and society. Examples range from economics to medicine, from biology to physics, and from social to computer sciences. Repetitions or recurrences of similar phenomena are a fundamental characteristic of non-random natural and artificial systems and, as a measurement of the activity of such systems, time series often include pairs of segments of strikingly high similarity. These segment pairs are commonly called motifs~\cite{Lin02WTDM}, and their existence is unlikely to be due to chance alone. In fact, they usually carry important information about the underlying system. Thus, motif discovery is fundamental for understanding, characterizing, modeling, and predicting the system behind the time series~\cite{Mueen14WIRES}. Besides, motif discovery is a core part of several higher-level algorithms dealing with time series, in particular classification, clustering, summarization, compression, and rule-discovery algorithms (see, e.g., references in~\cite{Mueen14WIRES,Mueen13ICDM}).

Identifying similar segment pairs or motifs implies examining all pairwise comparisons between all possible segments in a time series. This, specially when dealing with long time series streams, results in prohibitive time and space complexities. It is for this reason that the majority of motif discovery algorithms resort to some kind of data discretization or approximation that allows them to hash and retrieve segments efficiently. Following the works by Lin~et~al.~\cite{Lin02WTDM} and Chiu~et~al.~\cite{Chiu03KDD}, many of such approaches employ the SAX representation~\cite{Lin07DMKD} and/or a sparse collision matrix~\cite{Buhler02JCB}. These allow them to achieve a theoretically low computational complexity, but sometimes at the expense of very high constant factors. In addition, approximate algorithms usually suffer from a number of data-dependent parameters that, in most situations, are not intuitive to set (e.g.,~time/amplitude resolutions, dissimilarity radius, segment length, minimum segment frequency, etc.). 

A few recent approaches overcome some of these limitations. For instance, Castro~\&~Azevedo~\cite{Castro10SDM} propose an amplitude multi-resolution approach to detect frequent segments, Li~\&~Lin~\cite{Li10MDM} use a grammar inference algorithm for exploring motifs with lengths above a certain threshold, Wilson et al.~\cite{Wilson08IJAC} use concepts from immune memory to deal with different lengths, and Floratou~et~al.~\cite{Floratou11TKDE} combine suffix trees with segment models to find motifs of any length. Nevertheless, in general, these approaches still suffer from other data-dependent parameters whose correct tuning can require considerable time. In addition, approximate algorithms are restricted to a specific dissimilarity measure between segments (the one implicit in their discretization step) and do not allow easy access to preliminary results, which is commonly known as anytime algorithms~\cite{Zilberstein96AIM}. Finally, to the best of our knowledge, only~\cite{Tanaka05ML,Yankov07KDD,Tang08KBS} consider the identification of motif pairs containing segments of different lengths. This can be considered a relevant feature, as it produces better results in a number of different domains~\cite{Yankov07KDD}.

In contrast to approximate approaches, algorithms that do not discretize the data have been comparatively much less popular, with low efficiency generally. Exceptions to this statement achieved efficiency by sampling the data stream~\cite{Catalano06PKDD} or by identifying extreme points that constrained the search~\cite{Mohammad09NGC}. In fact, until the work of Mueen~et~al.~\cite{Mueen09SDM}, the exact identification of time series motifs was thought to be intractable for even time series of moderate length. In said work, a clever segment ordering was combined with a lower bound based on the triangular inequality to yield the true, exact, most similar motif. According to the authors, the proposed algorithm was more efficient than existing approaches, including all exact and many approximate ones~\cite{Mueen09SDM}. After Mueen~et~al.'s work, a number of improvements have been proposed, the majority focusing on eliminating the need to set a fixed segment length~\cite{Nunthanid11ECTI,Yingcharxxx13ICDM,Mohammad14ACIIDS}.

Mueen himself has recently published a variable-length motif discovery algorithm which clearly outperforms the iterative search for the optimal length using~\cite{Mueen09SDM} and, from the reported numbers, also outperforms further approaches such as~\cite{Nunthanid11ECTI,Yingcharxxx13ICDM,Mohammad14ACIIDS}. This algorithm, called MOEN~\cite{Mueen13ICDM}, is essentially parameter-free, and is believed to be one of the most efficient motif discovery algorithms available nowadays. However, its complexity is still quadratic in the length of the time series~\cite{Mueen13ICDM}, and hence its applicability to large-scale time series streams remains problematic. Furthermore, in order to derive the lower bounds used, the algorithm is restricted with regard to the dissimilarity measure used to compare time series segments (Euclidean distance after z-normalization). In general, exact motif discovery algorithms have important restrictions with regard to the dissimilarity measure, and many of them still suffer from being non-intuitive and tedious to tune parameters. Moreover, few of them allow for anytime versions and, to the best of our knowledge, not one of them is able to identify motif pairs containing segments of different lengths.

In this article, we propose a new standpoint to time series motif discovery by treating the problem as an anytime multimodal optimization task. To the best of our knowledge, this standpoint is completely unseen in the literature. Here, we firstly reason and discuss its multiple advantages (Sec.~\ref{sec:motif}). Next, we present SWARMMOTIF (Sec.~\ref{sec:swarmmotif}), an anytime algorithm for time series motif discovery based on particle swarm optimization (PSO). We subsequently evaluate the performance of the proposed approach using 9~different real-world time series from distinct domains (Sec.~\ref{sec:eval}). Our results show that SWARMMOTIF is extremely competitive when compared to the state-of-the-art, obtaining motif pairs of comparable similarity in considerably less time and with minimum storage requirements (Sec.~\ref{sec:results}). Moreover, we show that SWARMMOTIF is significantly robust against different implementation choices. 
To conclude, we briefly comment on the application of multimodal optimization techniques to time series analysis and mining, which we believe has great potential (Sec.~\ref{sec:conclusion}). The data and code used in our experiments will available online
. 

\section{Time Series Motif Discovery as an Anytime Multimodal Optimization Task}
\label{sec:motif}

\subsection{Definitions and Task Complexity}
\label{sec:motifdefs}

\begin{figure}[t]
	\centering
	\includegraphics{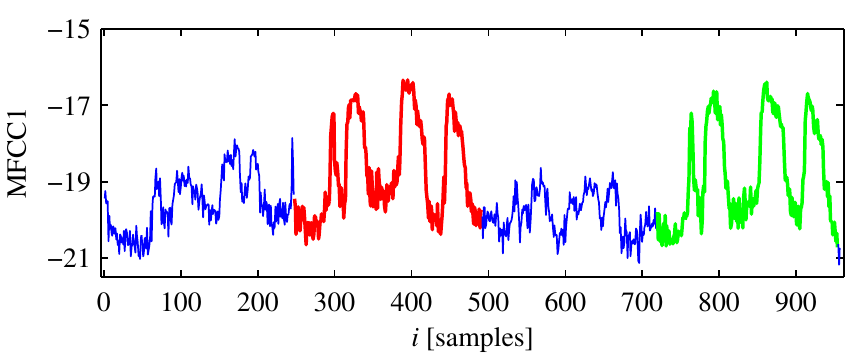}
	\caption{Example of a time series motif pair found in the \textsc{Wildlife} time series of~\cite{Mueen10KDD} using SWARMMOTIF and normalized dynamic time warping as the dissimilarity measure: $a=248$, $w_a=244$, $b=720$, and $w_b=235$. Note that $w_a\neq w_b$ and, hence, a warping of the two segments needs to take place. This specific solution cannot be found by any of the approaches mentioned in Sec.~\ref{sec:intro}.}
	\label{fig:example}
\end{figure}

From the work by Mueen et al.~\cite{Mueen09SDM,Mueen13ICDM}, we can derive a formal, generic similarity-based definition~\cite{Mueen14WIRES} of time series motifs. Given a time series $\textbf{z}$ of length $n$, $\textbf{z}=[z_1,\dots z_n]$, a normalized segment dissimilarity measure $D$, and a temporal window of interest between $w_{\text{min}}$ and $w_{\text{max}}$ samples, the top-$k$ time series motifs $\mathcal{M}=\{\textbf{m}_1,\dots \textbf{m}_k\}$ correspond to the $k$ most similar segment pairs $\textbf{z}_a^{w_a}=[z_a,\dots z_{a+w_a-1}]$ and $\textbf{z}_b^{w_b}=[z_b,\dots z_{b+w_b-1}]$, for $w_a,w_b\in[w_{\text{min}},w_{\text{max}}]$, $a\in[1,n-w_a+1]$, and $b\in[1,n-w_b+1]$ where, in order to avoid repeated and trivial matches~\cite{Lin02WTDM}, $a+w_a< b$. Thus, the $i$-th motif can be fully described by the tuple $\textbf{m}_i=\{a,w_a,b,w_b\}$. The motifs in $\mathcal{M}$ are non-overlapping\footnote{Notice that, following~\cite{Mueen13ICDM}, this definition can be trivially extended to different degrees of overlap.} and ordered from lowest to highest dissimilarity such that $D(\textbf{m}_1)\leq D(\textbf{m}_2) \leq \dots \leq D(\textbf{m}_k)$ where $D(\textbf{m}_i)=D(\{a,w_a,b,w_b\})=D(\textbf{z}_a^{w_a},\textbf{z}_b^{w_b})$. An example of a time series motif pair from a real data set is shown in Fig.~\ref{fig:example}.

It is important to stress that $D$ needs to normalize with respect to the lengths of the considered segments. Otherwise, we would not be able to compare motifs of different lengths. There are many ways to normalize with respect to the length of the considered segments. Ratanamahatana~\&~Keogh~\cite{Ratanamahatana04WMTSD} list a number of intuitive normalization mechanisms for dynamic time warping that can easily be applied to other measures. For instance, in the case of a dissimilarity measure based on the $\text{L}_{\text{p}}$ norm~\cite{Serra14KBS}, we can directly divide by the segment length\footnote{The only exception is with  $\text{L}_{\infty}$, which could be considered as already being normalized.}, using brute-force upsampling to the largest length~\cite{Ratanamahatana04WMTSD} when $w_a\neq w_b$. 

From the definitions above, we can see that a brute-force search in the motif space for the most similar motifs is of $O(n^2{w_\Delta}^2)$, where $w_\Delta=w_{\text{max}}-w_{\text{min}}+1$ (for the final time complexity one needs to further multiply by the cost of calculating $D$). Hence, for instance, in a perfectly feasible case where $n=10^7$ and $w_\Delta=10^3$, we have $10^{20}$ possibilities. Magnitudes like this challenge the memory and speed of any optimization algorithm, specially if we have no clue to guide the search~\cite{Hillinger10BOOK}. However, it is one of our main objectives to show here that time series generally provide some continuity to this search space, and that this continuity can be exploited by optimization algorithms.

\subsection{Continuity}
\label{sec:motifcont}

\begin{figure}[t]
	\centering
	\includegraphics{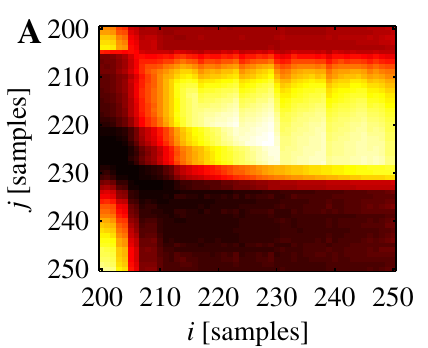}
	\includegraphics{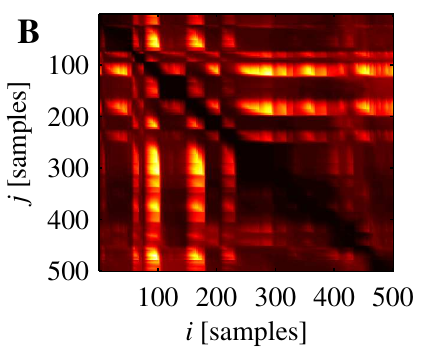}
	\caption{Visualizations of the search space obtained with the \textsc{Wildlife} time series~\cite{Mueen10KDD} and dynamic time warping as the dissimilarity measure: (A) fixing $a=110$ and $b=602$ but $i,j=200,\dots 250$ and (B) fixing $w_a=222$ and $w_b=240$ but $i,j=1,\dots 500$. Darker colors corresponds to more similarity.}
	\label{fig:continuity}
\end{figure}

A fundamental property of time series is autocorrelation, implying that consecutive samples in a time series have some degree of resemblance and that, most of the time, we do not observe extremal differences between them\footnote{If a time series had no autocorrelation, we might better treat it as an independent random process.}. This property, together with the established ways of computing similarity between time series~\cite{Serra14KBS}, is what gives continuity to our search space. Consider a typical dissimilarity measure like dynamic time warping between z-normalized segments and the time series of Fig.~\ref{fig:example}. If we fix the motif starting points $a$ and $b$ to some random values, we can compute $D(\textbf{z}_a^{i},\textbf{z}_b^{j})$ for $i,j=w_{\text{min}},\dots, w_{\text{max}}$ (Fig.~\ref{fig:continuity}A). We see that these two dimensions have a clear continuity, i.e.,~that $D(\textbf{z}_a^{i},\textbf{z}_b^{j}) \sim D(\textbf{z}_a^{i+1},\textbf{z}_b^{j}) \sim D(\textbf{z}_a^{i},\textbf{z}_b^{j+1}) \sim D(\textbf{z}_a^{i+1},\textbf{z}_b^{j+1})$, and so forth. Similarly, if we fix the motif lengths $w_a$ and $w_b$ to some random values, we can compute $D(\textbf{z}_i^{w_a},\textbf{z}_j^{w_b})$ for $i=1,\dots n-w_a$ and $j=1,\dots n-w_b$ (Fig.~\ref{fig:continuity}B). We see that the remaining two dimensions of the problem also have some continuity, i.e.,~$D(\textbf{z}_i^{w_a},\textbf{z}_j^{w_b+j}) \sim D(\textbf{z}_{i+1}^{w_a},\textbf{z}_{j}^{w_b}) \sim D(\textbf{z}_{i}^{w_a},\textbf{z}_{j+1}^{w_b}) \sim D(\textbf{z}_{i+1}^{w_a},\textbf{z}_{j+1}^{w_b})$, and so forth. The result is a four-dimensional, multimodal, continuous but noisy\footnote{We use the term noisy here to stress that the continuity of the space may be altered at some points due to potential noise in the time series. It is not the case that we have a noisy, unreliable dissimilarity measurement $D$ that could change in successive evaluations.} motif space, where the dissimilarity $D$ acts as the fitness measure and the top-$k$ valley peaks (considering dissimilarity) correspond to the top-$k$ motifs in $\mathcal{M}$.

\subsection{Anytime Solutions}
\label{sec:motifany}

Finding an optimization algorithm that can locate the global minima of the previous search spaces faster than existing motif discovery algorithms can be a difficult task. However, we have robust and established algorithms for efficiently locating prominent local minima in complex search spaces~\cite{Blum03ACMCS,Jin05TEVC,Bianchi09NC}. Hence, we can intuitively devise a simple strategy: if we keep the best found minima and randomly reinitialize the optimization algorithm every time it stagnates, we should, sooner or later, start locating the global minima. In the meantime, we could have obtained relatively good candidates. This corresponds to the basic paradigm of anytime algorithms~\cite{Zilberstein96AIM}.

\begin{figure}[t]
	\centering
	\includegraphics{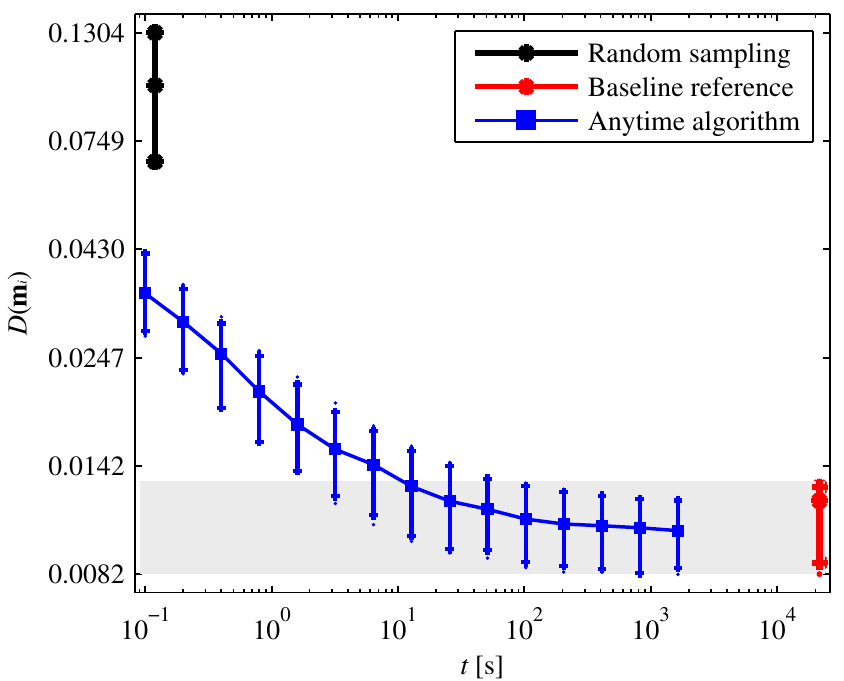}
	\caption{Schema of a plot to assess the performance of an anytime motif discovery algorithm (blue curve). Error bars indicate 5~and 95~percentiles of $D(\textbf{m}_i)$, their central marker indicates the median, and the isolated dots indicate maximum and minimum values. The gray area at the bottom denotes the area where $D(\textbf{m}_i)^\ast$ lie. The top-left black error bar acts as a reference and shows the range of $D(\textbf{m}_i)$ obtained by random sampling the motif space. The bottom-right red error bar is placed at the time that the baseline algorithm spent in the calculations. Better performing anytime algorithms have a curve closer to the bottom left corner, quickly entering the gray area as their execution time increases. Notice that both axes are logarithmic.}
	\label{fig:anytime}
\end{figure}

Anytime algorithms have recently been highlighted as ``very beneficial for motif discovery in massive [time series] datasets''~\cite{Yingcharxxx13ICDM}. In an anytime algorithm for motif discovery, $D(\textbf{m}_i)$ improves over time, until it reaches the top-$k$ dissimilarity values $D(\textbf{m}_i)^\ast$ obtained by a brute-force search approach. Thus, we gradually improve $\mathcal{M}$ until we reach the true exact solution $\mathcal{M}^\ast$. A good anytime algorithm will quickly find low $D(\textbf{m}_i)$, ideally reaching $D(\textbf{m}_i)^\ast$ earlier than its non-anytime competitors (Fig.~\ref{fig:anytime}). 

Note that a sufficiently good $\mathcal{M}$ may suffice in most situations, without the need that $\mathcal{M}=\mathcal{M}^\ast$. This is particularly true for more exploratory tasks, where one is typically interested in data understanding and visual inspection (see~\cite{Mueen14WIRES}), and can also hold for other tasks, as top-$k$ motifs can be very similar among themselves. In the latter situation, given a seed within $\mathcal{M}^\ast$, we can easily and efficiently retrieve further repetitions via common established approaches~\cite{Kahveci04TKDE,Rakthanmanon12KDD}. Thus, only non-frequent or singular motifs may be missed. These can be valuable too, as the fact that they are non-frequent does not imply that they cannot carry important information (think for example of extreme events of interest that perhaps only happen twice in a measurement). For those singular motifs, we can wait longer if using an anytime algorithm, or we can resort to the state-of-the-art if that is able to provide its output within an affordable time limit.

\subsection{Particle Swarm Optimization}
\label{sec:motifpso}

The continuity and anytime observations above (Secs.~\ref{sec:motifcont} and~\ref{sec:motifany}) relax the requirements for the optimization algorithm to be employed in the considered motif spaces. In fact, if we do not have to assess the global optimality of a solution, we have a number of approaches that can deal with large, multimodal, continuous but noisy search spaces~\cite{Blum03ACMCS,Jin05TEVC,Bianchi09NC}. Among them, we choose PSO~\cite{Clerc06BOOK,Poli07SI,Banks07aNC,Blum08BOOK,Parsopoulos10BOOK}. PSO is a population-based stochastic approach for solving continuous and discrete optimization problems~\cite{Blum08BOOK} which has been applied to multimodal problems~\cite{Barrera09BOOKCHAP}. It is a metaheuristic~\cite{Bianchi09NC}, meaning that it cannot guarantee whether the found solution corresponds to a global optimum. The original PSO algorithm cannot even guarantee the convergence to a local optimum, but adapted versions of it have been proven to solve this issue~\cite{VanDerBerg10FI}. Other versions guarantee the convergence to the global optimum, but only with the number of iterations approaching infinity~\cite{VanDerBerg10FI}.

PSO has gained increasing popularity among researchers and practitioners as a robust and efficient technique for solving difficult optimization problems. It makes few or no assumptions about the problem being optimized, does not require it to be differentiable, can search very large spaces of candidate solutions, and can be applied to problems that are irregular, incomplete, noisy, dynamic, etc.\ (see~\cite{Blum08BOOK, Clerc06BOOK, Poli07SI, Banks07aNC, Parsopoulos10BOOK, Barrera09BOOKCHAP} and references therein). PSO iteratively tries to improve a candidate solution with regard to a given measure of quality or fitness function. Hence, furthermore, it can be considered an anytime algorithm.

\subsection{Advantages of an Optimization-Based Solution Using Particle Swarms}
\label{sec:motifadv}

Notice that treating time series motif discovery as an optimization problem naturally yields several advantages:
\begin{enumerate}
	\item We do not require much memory, as we can basically store only the stream time series and preprocess the required segments at every fitness evaluation.
	\item We are able to achieve a certain efficiency, as optimization algorithms do not usually explore the full solution space and perform few fitness evaluations~\cite{Hillinger10BOOK}.
	\item We can employ any dissimilarity measure $D$ as our fitness function. Its only requirements are segment length independence and a minimal search space continuity. Intuitively, this holds for the high majority of time series dissimilarity measures that are currently used (Secs.~\ref{sec:motifdefs} and~\ref{sec:motifcont}). Additionally, we can straightforwardly incorporate notions of `interestingness', hubness, or complexity (see references in~\cite{Serra14KBS}). This flexibility is very uncommon in current time series motif discovery algorithms (Sec.~\ref{sec:intro}).
	\item We do not need to force the two segments of the motif to be of the same length. The dissimilarity function $D$ can expressly handle segments of different lengths or we can simply upsample to the largest length (see~\cite{Ratanamahatana04WMTSD}). Although considering different segment lengths has been highlighted as an objectively better approach, practically none of the current time series motif discovery algorithms contemplates this option (Sec.~\ref{sec:intro}).
	\item Since we search for the optimal $w_a$ and $w_b$, together with $a$ and $b$, we do not need to set the exact segment lengths as a parameter. Instead, we can use a more intuitive and easier to set range of lengths $w_a,w_b\in[w_{\text{min}},w_{\text{max}}]$.
	\item We can easily modify our fitness criterion to work with different task settings. Thus, just by replacing $D$, we are able to work with multi-dimensional time series~\cite{Hao14IV}, detect sub-dimensional motifs~\cite{Minnen07ICDM}, perform a constrained motif discovery task~\cite{Mohammad09NGC}, etc.
	\item We can incorporate notions of motif frequency to our fitness function and hence expand our similarity-based definition of motif to incorporate both notions~\cite{Mueen14WIRES}. For instance, instead of optimizing for individual motifs $\textbf{m}_i$, we can optimize sets of motifs $\mathcal{M}'_i$ of size $r_i$ such that $\frac{1}{r_i} \sum_{\textbf{m}_j\in \mathcal{M}'_i} D(\textbf{m}_j)$ is minimal. We can choose $r_i$ to be a minimum frequency of motif appearance or we can even decide to optimize it following any suitable criterion.
\end{enumerate}
In addition, using PSO has a number of interesting properties, some of which may be shared with other metaheurisics:
\begin{enumerate}
	\item We have a straightforward mapping to the problem at hand (Sec.~\ref{sec:swarmmotifmain}).
	\item By construction, we have an anytime algorithm (Sec.~\ref{sec:motifpso}).
	\item We can obtain accurate and much faster solutions, as compared to the state-of-the-art in time series motif discovery (Sec.~\ref{sec:resfinal}).
	\item We have an essentially parameter-free algorithm~\cite{Blum08BOOK}. As will be shown, all our parameter choices turn out to be non-critical to achieve the most competitive performances (Secs.~\ref{sec:resparam} and~\ref{sec:resvariants}).
	\item We have an easily parallelizable algorithm. The agent-based nature of PSO naturally yields to parallel implementations~\cite{Banks07aNC}.
	\item We still have the possibility to apply lower bounding techniques to $D$ in order to reduce its computational cost~\cite{Mueen14WIRES,Rakthanmanon12KDD}. Among others, we may exploit the particles' best-so-far values or spatially close dissimilarities.
	\item All of these use a simple, easy to implement algorithm requiring low storage capabilities (Sec.~\ref{sec:swarmmotifdetail}).
\end{enumerate}

\section{SWARMMOTIF}
\label{sec:swarmmotif}

\subsection{Main Algorithm}
\label{sec:swarmmotifmain}

Our PSO approach to time series motif discovery is based on the combination of two well-known extensions to the canonical PSO~\cite{Poli07SI}. On one hand, we employ multiple reinitializations of the swarm on stagnation~\cite{Eberhart01CEC}. On the other hand, we exploit the particles' ``local memories'' with the intention of forming stable niches across different local minima~\cite{Li10TEVC}. The former emulates a parallel multi-swarm approach~\cite{Barrera09BOOKCHAP} without the need of having to define the number of swarms and their communication. The latter, when combined with the former, results in a low-complexity niching strategy~\cite{Barrera09BOOKCHAP} that does not require niching parameters (see the related discussion in~\cite{Bonyadi14SEC,Bonyadi14BOOKCHAP}). SWARMMOTIF, the implementation of the two extensions, is detailed in Algorithm~\ref{alg:pso}.

\begin{algorithm}[t]
	\caption{\textsc{SWARMMOTIF}($\textbf{z}$,$D$,$w_{\text{min}}$,$w_{\text{max}}$,$k$,$t_{\text{max}}$)}
	\label{alg:pso}
	\algsetup{indent=0.5cm}
	\begin{algorithmic}[1]
	\vspace*{0.1cm}
	\REQUIRE Time series $\textbf{z}$ of length $n$, dissimilarity measure $D$, minimum and maximum segment length $w_{\text{min}}$ and $w_{\text{max}}$, number of motifs $k$, and maximum amount of time (number of iterations) $t_{\text{max}}$. \\
	{\renewcommand{\algorithmicrequire}{\textbf{Require:}}
	\REQUIRE Number of particles $\kappa$, topology $\theta$, constriction constant $\phi$, and maximum amount of time at stagnation (number of iterations) $\tau$. \\
	}
	\ENSURE A set of motifs $\mathcal{M}$. \\
	\vspace*{0.1cm}
	\STATE $c_0,c_1,c_2 \leftarrow$ \textsc{GetConstants}($\phi$)
	\STATE $\mathcal{X},\mathcal{V},\mathcal{S},\mathcal{P} \leftarrow$ \textsc{InitializeSwarm}($n$,$w_{\text{min}}$,$w_{\text{max}}$,$\kappa$)
	\STATE $\Theta \leftarrow$ \textsc{InitializeTopology}($\theta$,$\kappa$)
	\STATE $s^\ast \leftarrow \infty$
	\STATE $\mathcal{M} \leftarrow$ \textsc{EmptyPriorityQueue}()
	\FOR{$t=1,\dots t_{\text{max}}$}
		\FOR{$i=1,\dots \kappa$}
			\IF{ \textsc{ValidPosition}($\textbf{x}_i$)}
				\STATE $d \leftarrow D(\textbf{x}_i)$
				\IF{$d < s_i$}
					\STATE $s_i \leftarrow d$
					\STATE $\textbf{p}_i \leftarrow \textbf{x}_i$
					\STATE $\mathcal{M}$.\textsc{Push}($d$,$\textbf{x}_i$)
					\IF{$d < s^\ast$}
						\STATE $s^\ast \leftarrow d$
						\STATE $t_{\text{update}} \leftarrow t$
					\ENDIF
				\ENDIF
			\ENDIF
		\ENDFOR
		\FOR{$i=1,\dots \kappa$}
			\STATE $g \leftarrow i$
			\FOR{$j$ \textbf{in} $\Theta_i$}
				\IF{$s_j \leq s_g$}
					\STATE $g \leftarrow j$
				\ENDIF
			\ENDFOR
			\STATE $\textbf{v}_i \leftarrow c_0\textbf{v}_i + c_1\textbf{u}_1\otimes(\textbf{p}_i-\textbf{x}_i) + c_2\textbf{u}_2\otimes(\textbf{p}_g-\textbf{x}_i)$
			\STATE $\textbf{x}_i \leftarrow \textbf{x}_i + \textbf{v}_i$
		\ENDFOR
		\IF{$t-t_{\text{update}} = \tau$}
			\STATE $\mathcal{X},\mathcal{V},\mathcal{S},\mathcal{P} \leftarrow$ \textsc{InitializeSwarm}($n$,$w_{\text{min}}$,$w_{\text{max}}$,$\kappa$)
			\STATE $s^\ast \leftarrow \infty$
		\ENDIF
	\ENDFOR
	\RETURN \textsc{NonOverlapping}($\mathcal{M}$,$k$) \\
	\end{algorithmic}
\end{algorithm}

SWARMMOTIF takes a time series $\textbf{z}$ of length $n$ as input, together with a segment dissimilarity measure $D$, and the range of segment lengths of interest, limited by $w_{\text{min}}$ and $w_{\text{max}}$. The user also needs to specify $k$, the desired number of motifs, and $t_{\text{max}}$, the maximum time spent by the algorithm (in iterations\footnote{The number of iterations is easy to infer from the available time as, for the same input, the elapsed time will be roughly directly proportional to the number of iterations.}). SWARMMOTIF outputs a set of $k$ non-overlapping motifs $\mathcal{M}$. We implement $\mathcal{M}$ as a priority queue, which typically stores more than $k$ elements to ensure that it contains $k$ non-overlapping motifs. This way, by sorting the motif candidates as soon as they are found, we allow potential queries to $\mathcal{M}$ at any time during the algorithm's execution. In that case, we only need to dynamically check the candidates' overlap (Sec.~\ref{sec:swarmmotifdetail}). Notice that $n$, $D$, $w_{\text{min}}$, $w_{\text{max}}$, $k$, and $t_{\text{max}}$ are not parameters of the algorithm, but requirements of the task (they depend on the data, the problem, and the available time). The only parameters to be set, as specified in  Algorithm~\ref{alg:pso}'s requirements, are the number of particles $\kappa$, the topology $\theta$, the constriction constant $\phi$, and the maximum amount of iterations at stagnation $\tau$. Nevertheless, we will show that practically none of the possible parameter choices introduces a significant variation in the reported performance (Sec.~\ref{sec:resparam}).

Having clarified SWARMMOTIF's input, output, and requirements, we now elaborate on its procedures. Algorithm~\ref{alg:pso} starts by computing the velocity update constants (line~1) following Clerc's constriction method~\cite{Clerc02TEVC}, i.e.,
\begin{equation*}
	c_0 = \frac{2}{\left\vert 2-\phi-\sqrt{\phi^2-4\phi} \right\vert}
\end{equation*}
and
\begin{equation}
	c_1=c_2=c_0\phi/2 .
	\label{eq:c1c2}
\end{equation}
Next, a swarm with $\kappa$ particles is initialized (line~2). The swarm is formed by four data structures: a set of particle positions $\mathcal{X}=\{\textbf{x}_1,\dots \textbf{x}_\kappa\}$, a set of particle velocities $\mathcal{V}=\{\textbf{v}_1,\dots \textbf{v}_\kappa\}$, a set of particle best scores $\mathcal{S}=\{s_1,\dots s_\kappa\}$, and a set of particle best positions $\mathcal{P}=\{\textbf{p}_1,\dots \textbf{p}_\kappa\}$ (the initialization of these four data structures is detailed in Algorithm~\ref{alg:swarminit}). Particles' positions $\textbf{x}_i$ and $\textbf{p}_i$ completely determine a motif candidate, and have a direct correspondence with $\textbf{m}_i$ (see Sec.~\ref{sec:swarmmotifdetail}). A further data structure $\Theta$ indicates the indices of the neighbors of each particle according to a given social topology $\theta$ (line~3). Apart from the swarm, we also initialize a global best score $s^\ast$ (line~4) and the priority queue $\mathcal{M}$ (line~5). We then enter the main loop (lines~6--26). In it, we perform three main actions. Firstly, we compute the particles' fitness and perform the necessary updates (lines~7--16). Secondly, we modify the particles' position and velocity using their personal and neighborhood best positions (lines~17--23). Thirdly, we control for stagnation and reinitialize the swarm if needed (lines~24--26). Finally, when we exit the loop, we return the first $k$ non-overlapping motif candidates from $\mathcal{M}$ (line~27).

\begin{algorithm}[t]
	\caption{\textsc{InitializeSwarm}($n$,$w_{\text{min}}$,$w_{\text{max}}$,$\kappa$)}
	\label{alg:swarminit}
	\algsetup{indent=0.5cm}
	\begin{algorithmic}[1]
	\vspace*{0.1cm}
	\REQUIRE Time series length $n$, minimum and maximum segment length $w_{\text{min}}$ and $w_{\text{max}}$, and number of particles $\kappa$. \\
	\ENSURE Particle positions $\mathcal{X}$, velocities $\mathcal{V}$, best scores $\mathcal{S}$, and best positions $\mathcal{P}$. \\
	\vspace*{0.1cm}
	\FOR{$i=1,\dots \kappa$}
		\STATE $x_{i,2} \leftarrow w_{\text{min}}+(w_{\text{max}}+1-w_{\text{min}})u$
		\STATE $x_{i,4} \leftarrow w_{\text{min}}+(w_{\text{max}}+1-w_{\text{min}})u$
		\STATE $x_{i,1} \leftarrow 1+(n-x_{i,2})(1-\sqrt{u})$
		\STATE $x_{i,3} \leftarrow 1+(n-x_{i,4}-(x_{i,1}+x_{i,2}))u$
		\STATE $ \textbf{x}' \leftarrow$ \textit{As in lines 2--5}
		\STATE $\textbf{v}_i \leftarrow \textbf{x}'-\textbf{x}_i$
		\STATE $s_i \leftarrow \infty$
		\STATE $\textbf{p}_i \leftarrow \textbf{x}_i$
	\ENDFOR
	\RETURN $\mathcal{X},\mathcal{V},\mathcal{S},\mathcal{P}$ \\
	\end{algorithmic}
\end{algorithm}

The particles' fitness loop (lines~7--16) can be described as follows. For the particles that have a valid position within the ranges used for particle initializations (line~8; see also Algorithm~\ref{alg:swarminit} for initializations), we calculate their fitness $D$ (line~9) and, if needed, update their personal bests $s_i$ and $\textbf{p}_i$ (lines 10--12). As mentioned, $D$ needs to be independent of the segments' lengths, which is typically an easy condition for time series dissimilarity measures (Sec.~\ref{sec:motifdefs}). In the case that the particles find a new personal best, we save the motif dissimilarity $d$ and its position $\textbf{x}_i$ into $\mathcal{M}$ (line~13). Next, we update $t_{\text{update}}$, the last iteration when an improvement of the global best score $s^\ast$ has occurred (lines~14--16).

The particles' update loop (lines~17--23) is straightforward. We first select each particle's best neighbor $g$ using the neighborhood personal best scores $s_j$ (lines~18--21). Then, we use the positions of the best neighbor's personal best $\textbf{p}_g$ and the particle's personal best $\textbf{p}_i$ to compute its new velocity and position (lines~22--23). We employ component-wise multiplication, denoted by $\otimes$, and two random vectors $\textbf{u}_1$ and $\textbf{u}_2$ whose individual components $u_{i,j}=U(0,1)$, being $U(l,h)$ a uniform real random number generator such that $l\leq U(l,h) < h$. Note that by considering the particles' neighborhood personal bests $\textbf{p}_g$ we follow the aforementioned local neighborhood niching strategy~\cite{Li10TEVC}. At the end of the loop we control for stagnation by counting the number of iterations since the last global best update and applying a threshold $\tau$ (line~24). Note that this is the mechanism responsible for the aforementioned multiple reinitialization strategy~\cite{Eberhart01CEC}.

The initialization of the swarm used in Algorithm~\ref{alg:pso} (lines~2 and~25) is further detailed in Algorithm~\ref{alg:swarminit}. In it, for each particle, two random positions $\textbf{x}_i$ and $\textbf{x}'$ are drawn (lines 2--6) and the initial velocity is computed as the subtraction of the two (line 7). To obtain $\textbf{x}_i$ and $\textbf{x}'$, uniform real random numbers $u=U(0,1)$ are subsequently generated. The personal best score $s_i$ is set to infinite (line 8) and $\textbf{x}_i$ is taken as the current best position $\textbf{p}_i$ (line 9). Note that $\sqrt{u}$ (line 4) is used to ensure a uniform distribution of the particles across the triangular subspace formed by $x_{i,1}$ and $x_{i,3}$ (line 5; see also Sec.~\ref{sec:motifdefs}).

\subsection{Implementation Details}
\label{sec:swarmmotifdetail}

Some implementation details are missing in Algorithms~\ref{alg:pso} and~\ref{alg:swarminit}. Firstly, positions $\textbf{x}_i$ are floored component-wise inside \textsc{ValidPosition}, $D$, and $\mathcal{M}$ (thus obtaining motif $\textbf{m}_i$). Secondly, the motif priority queue $\mathcal{M}$ is implemented as an associative container (logarithmic insertion time) that sorts its elements according to $d$ and stores $\textbf{m}_i$. Thirdly, the last visited positions are cached into a hash table (constant lookup time) in order to avoid some of the possible repeated dissimilarity computations. Fourthly, we incorporate the option to constrain the motif search by specifying a maximum segment stretch in Algorithm~\ref{alg:swarminit} and \textsc{ValidPosition}. Finally, the function that returns the non-overlapping top-$k$ motifs employs a boolean array of size $n$ in order to avoid $O(k^2)$ comparisons between members of the queue (cf.~\cite{Mueen13ICDM}). Notice that we have a memory-efficient implementation, as we basically only need to store $\textbf{z}$ and the boolean array (both of $O(n)$ space), $\mathcal{M}$ (of $O(k)$ space, $k\ll n$), and $\mathcal{X}$, $\mathcal{V}$, $\mathcal{S}$, $\mathcal{P}$, and $\Theta$ (all of them of $O(\kappa)$ space, $\kappa\ll n$). The aforementioned hash table (optional) can be allocated in any predefined, available memory segment. For the sake of brevity, the interested reader is referred to the provided code for a full account of the outlined implementation details.

\subsection{Variants}
\label{sec:swarmmotifvariants}

Given the main Algorithm~\ref{alg:pso}, we consider a number of variations that may potentially improve SWARMMOTIF's performance without introducing too much algorithmic complexity:
\begin{itemize}
	\item Sociability: We study whether a ``cognitive-only'' model, a ``social-only'' one, or different weightings of the two yield to some improvements~\cite{Kennedy97CEC}. To do so, we just need to introduce a parameter $\alpha\in[0,1]$ controlling the degree of `sociability' of the particles, and implement lines~1--2 of Algorithm~\ref{alg:variants} instead of Eq.~\ref{eq:c1c2}.
	\item Stochastic: We investigate the use of a random inertia weight~\cite{Eberhart01CEC}. This may alleviate the need of using the same $c_0$ in different environments, providing a potentially more adaptive trade-off between exploration and exploitation (also controlled by $\alpha$ in the previous point). To consider this variant, we just need to replace line~22 in Algorithm~\ref{alg:pso} by line~3 in Algorithm~\ref{alg:variants}.
	\item Velocity clamping: In addition to constriction, we study limiting the maximum velocity of the particles~\cite{Kennedy95ICNN}. Empirical studies have shown that the simultaneous consideration of a constriction factor and velocity clamping results in improved performance on certain problems~\cite{Eberhart00CEC}. To apply velocity clamping we add lines~4--6 of Algorithm~\ref{alg:variants} between lines~22 and~23 of Algorithm~\ref{alg:pso}. 
	\item Craziness: We introduce so-called ``craziness'' or ``perturbation'' in the particles' velocities, as initially suggested by Kennedy~\&~Eberhart~\cite{Kennedy95ICNN}. In such variant, inspired by the sudden direction changes observed in flocking birds, the particles' velocity is altered with a certain probability $\rho$, with the aim of favoring exploration by increasing directional diversity and discouraging premature convergence~\cite{Fourie00IWMDO}. We coincide with~\cite{Fourie00IWMDO} in that, in some sense, this can be seen as a mutation operation. To implement craziness we add lines 7--10 of Algorithm~\ref{alg:variants} between lines~22 and~23 of Algorithm~\ref{alg:pso}.
\end{itemize}

\begin{algorithm}[t]
	\caption{Variations to Algorithm~\ref{alg:pso}: sociability (lines~1--2), stochastic (line~3), velocity clamping (lines~4--6), and craziness (lines~7--10).}
	\label{alg:variants}
	\algsetup{indent=0.5cm}
	\begin{algorithmic}[1]
	\vspace*{0.1cm}
	\STATE $c_1 \leftarrow c_0\phi(1-\alpha)$
	\STATE $c_2 \leftarrow c_0\phi\alpha$
	\\ \vspace*{-0.1cm}
	\hrulefill
	\vspace*{0.1cm}
	\STATE $\textbf{v}_i \leftarrow (1-2(1-c_0))u\textbf{v}_i + c_1\textbf{u}_1\otimes(\textbf{p}_i-\textbf{x}_i) + c_2\textbf{u}_2\otimes(\textbf{p}_g-\textbf{x}_i)$
	\\ \vspace*{-0.1cm}
	\hrulefill
	\vspace*{0.1cm}
	\STATE $\textbf{v}^{\text{range}} \leftarrow [n,w_\Delta,n,w_\Delta]/2$
	\FOR{$v_j^{\text{range}}$ \textbf{in} $\textbf{v}^{\text{range}}$}
		\STATE $v_{i,j} \leftarrow \text{min}(v_j^{\text{range}},\text{max}(-v_j^{\text{range}},v_{i,j}))$
	\ENDFOR
	\vspace*{-0.1cm}
	\hrulefill
	\vspace*{0.1cm}
	\STATE $\textbf{v}'_i \leftarrow$ \textit{As in Algorithm~\ref{alg:swarminit}}
	\FOR{$v_{i,j}$ \textbf{in} $\textbf{v}_i$}
		\IF{$u<\rho$}
			\STATE $v_{i,j} \leftarrow v'_{i,j}$.
		\ENDIF
	\ENDFOR
	\end{algorithmic}
\end{algorithm}

\section{Evaluation Methodology}
\label{sec:eval}

To evaluate SWARMMOTIF's speed and accuracy we consider plots like the one presented in Fig.~\ref{fig:anytime}. As a reference, we draw uniform random samples from the motif search space and compute their dissimilarities (we take as many samples as the length $n$ of the time series). As a baseline, we use the top-25 motifs found by MOEN~\cite{Mueen13ICDM}, which we will denote by $\mathcal{M}^\ast$. Existing empirical evidence~\cite{Mueen09SDM,Mueen13ICDM} suggests that MOEN is the most efficient algorithm to retrieve the top exact similarity-based motifs in a range of lengths\footnote{Besides, we could not find any other promising exact or anytime approach with some available code, nor with sufficient detail to allow a reliable implementation.} (Sec.~\ref{sec:intro}). Notice furthermore that here we are not that interested in obtaining all top-25 true exact motifs, but more concerned on obtaining good seed motifs within these using an anytime approach (Sec.~\ref{sec:motifany}). 

As its competitors, MOEN has however some limitations (Sec.~\ref{sec:intro}). Thus, to fairly compare results, we have to apply some constrains to our algorithm. Since MOEN can only use the Euclidean distance between z-normalized segments, here we also adopt this formulation for $D$. In addition, as MOEN only considers pairs of segments of the same length (without resampling), we have to constrain SWARMMOTIF so that $x_{i,2}=x_{i,4}$. Therefore, the reported motif dissimilarities $D(\textbf{m}_i)$ correspond to the Euclidean distance between two z-normalized segments of the same length (we divide by the length of the segments to compare different segment lengths, Sec.~\ref{sec:motifdefs}). In the reported experiments, SWARMMOTIF is run 10 times with $k=10$. We stop its execution when we find 95\% of $D(\textbf{m}_i)$ within $\mathcal{M}^\ast$. This way, we assess the time taken to retrieve any 10 motifs from those with  at least 95\% confidence. All experiments are performed using a single core of an Intel\textsuperscript{\textregistered} Xeon\textsuperscript{\textregistered} CPU E5-2620 at 2.00 GHz.

To demonstrate that SWARMMOTIF is not biased towards a particular data source, time series length, or motif length, we consider 9~different time series of varying length, coming from distinct domains, and a number of arbitrary but source-consistent motif lengths (Table~\ref{tab:datasets}). As mentioned, we make these time series and our code available online (Sec.~\ref{sec:intro}). Four of the time series have been used for motif discovery in previous studies~\cite{Mueen09SDM,Mueen09ICDM}, while the other five are employed here for the first time for this task:
\begin{enumerate}
	\item \textsc{DowJones}: The daily closing values of the Dow Jones average in the USA from May 2, 1885 to April 22, 2014~\cite{Williamson12WEB}.
	\item \textsc{CarCount}: The number of cars measured for the Glendale on ramp for the 101~North freeway in Los Angeles, CA, USA~\cite{Ihler06KDD}. The measurement was carried out by the Freeway Performance Measurement System\footnote{http://pems.dot.ca.gov} and the data was retrieved from the UCI Machine Learning Repository~\cite{UCI13WEB}. Segments of missing values were manually interpolated or removed.
	\begin{table}[!t]
\renewcommand{\arraystretch}{1.1}
\caption{Characteristics of the time series used for evaluation, together with the considered segment length ranges.}
\label{tab:datasets}
\renewcommand{\tabcolsep}{4.5pt}
\centering
\begin{tabular}{l|ccccc}
\hline\hline
Name 				& Duration 	& Sample Rate	& $n$ 		& $w_{\text{min}}$ 	& $w_{\text{max}}$ \\
\hline
\textsc{DowJones}	& 129\,y		& 1\,d$^{-1}$	& 35503		& 30					& 300 \\
\textsc{CarCount}	& 175\,d		& 1/5\,min$^{-1}$ & 47534	& 270				& 320 \\
\textsc{Insect} 		& 32\,min 	& 38\,Hz			& 73929		& 300				& 500 \\
\textsc{EEG} 		& 1\,h 		& 50\,Hz			& 180214		& 150				& 250 \\
\textsc{FieldRecording} & 3\,h 	& 21.5\,Hz		& 226383		& 50					& 150 \\
\textsc{Wind}		& 4\,y 		& 1/6\,min$^{-1}$ & 369138	& 200				& 280 \\
\textsc{Power}		& 32\,months 	& 1/5\,min$^{-1}$ & 410608	& 250				& 620 \\
\textsc{EOG}	 		& 9\,h 		& 25\,Hz			& 809948		& 150				& 200 \\
\textsc{RandomWalk} 	& -		 	& -				& 1000000	& 100				& 250 \\
\hline\hline
\end{tabular}
\end{table}
	\item \textsc{Insect}: The electrical penetration graph of a beet leafhopper (\textit{circulifer tenellus})~\cite{Mueen09SDM}. The time series was retrieved from Mueen's website\footnote{\url{http://www.cs.ucr.edu/~mueen/MK}}.
	\item \textsc{EEG}: A one hour electroencephalogram (in $\mu$V) from a single channel in a sleeping patient~\cite{Mueen09SDM}. The time series was retrieved from Mueen's website\footnote{\url{http://www.cs.ucr.edu/~mueen/OnlineMotif}} and, according to the authors, was smoothed and filtered using domain-standard procedures.
	\item \textsc{FieldRecording}: The spectral centroid (in Hz) of a field recording retrieved from Freesound\footnote{\url{http://www.freesound.org/people/JeffWojo/sounds/121250}}~\cite{Font13ACMMM}. We used the mean of the stereo channels and the spectral centroid (linear frequency) Vamp SDK example plugin from Sonic Visualizer~\cite{Cannam10ACMMM}. We used a Hann window of 8192~samples at 44.1~KHz with 75\% overlap.	\item \textsc{Wind}: The wind speed (in m/s) registered in the buoy of Rincon del San Jose, TX, USA, between January 1, 2010 and April 11, 2014. The time series was retrieved from the Texas Coastal Ocean Observation Network website\footnote{\url{http://lighthouse.tamucc.edu/pq}}. Segments of missing values were manually interpolated or removed.
	\item \textsc{Power}: The electric power consumption (in KW) of an individual household\footnote{\url{http://archive.ics.uci.edu/ml/datasets/Individual+household+electric+power+consumption}}. The data was retrieved from the UCI Machine Learning Repository~\cite{UCI13WEB}. We took the global active power, removed missing values, and downsampled the original time series by a factor of 5 using averaging and 50\% overlap.
	\item \textsc{EOG}: An electrooculogram tracking the eye movements of a sleeping patient~\cite{Goldberger00CIRCULATION}. We took the downsampled time series~\cite{Mueen09ICDM} from Mueen's web page\footnote{\url{http://www.cs.ucr.edu/~mueen/DAME}}.
	\item \textsc{RandomWalk}: A random walk time series. This was artificially synthesized using $z_{i+1}=z_i+N(0,1)$ for $i=2,\dots n$ and $z_1=0$, where $N(0,1)$ is a real Gaussian random number generator with zero mean and unit variance.
\end{enumerate}

To assess the statistical significance of the differences between alternative parameter settings, we employ a two stage approach. First, we consider all settings at the same time and perform the Friedman's test~\cite{Hollander99BOOK}, which is a non-parametric statistical test used to detect differences in treatments across multiple test attempts. We use as inputs the median values for all settings for 25 equally-spaced time steps. In the case some difference between settings is detected (i.e.,~we reject the null hypothesis that the settings' performances come from the same distribution), we proceed to the second stage. In it, we perform all possible pairwise comparisons between settings using the Wilcoxon signed-rank test~\cite{Hollander99BOOK}, another non-parametric statistical hypothesis test used for comparing matched samples. To counteract the problem of multiple comparisons and control the so-called family-wise error rate, we employ the Holm-Bonferroni correction~\cite{Holm79SJS}. In all statistical tests, we consider a significance level of 0.01.

\section{Results}
\label{sec:results}

\subsection{Configuration}
\label{sec:resparam}

\begin{figure}[t]
	\centering
	\includegraphics{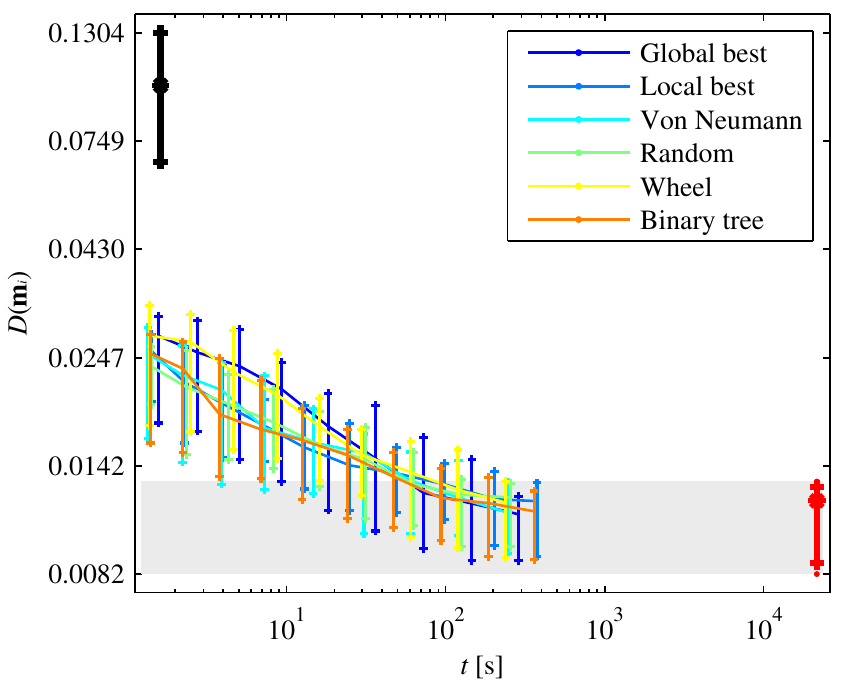}
	\caption{Effect of $\theta$ on the \textsc{EEG} time series. Equivalent results were observed with the other time series.}
	\label{fig:restopology}
\end{figure}

In pre-analysis, and according to common practice, we set $\kappa=100$, $\phi=4.1$, and $\tau=2000$. We then experimented with 6~different static topologies $\theta$~\cite{Mendes04THESIS}: global best, local best (two neighbors), Von Neumann, random (three neighbors), wheel, and binary tree. The results showed the qualitative equivalence of all topologies except, perhaps, global best and wheel (Fig.~\ref{fig:restopology}). In some data sets, these two turned out to yield slightly worse performances for short-time runs of the algorithm (small $t$), although for longer runs they gradually became equivalent to the rest. However, in general, no systematic statistically significant difference was detected between topologies. With this in mind, we chose the local best topology to further favor exploration and parallelism, and to be more consistent with our local neighborhood design principle of Sec.~\ref{sec:swarmmotifmain}. 

\begin{figure}[t]
	\centering
	\includegraphics{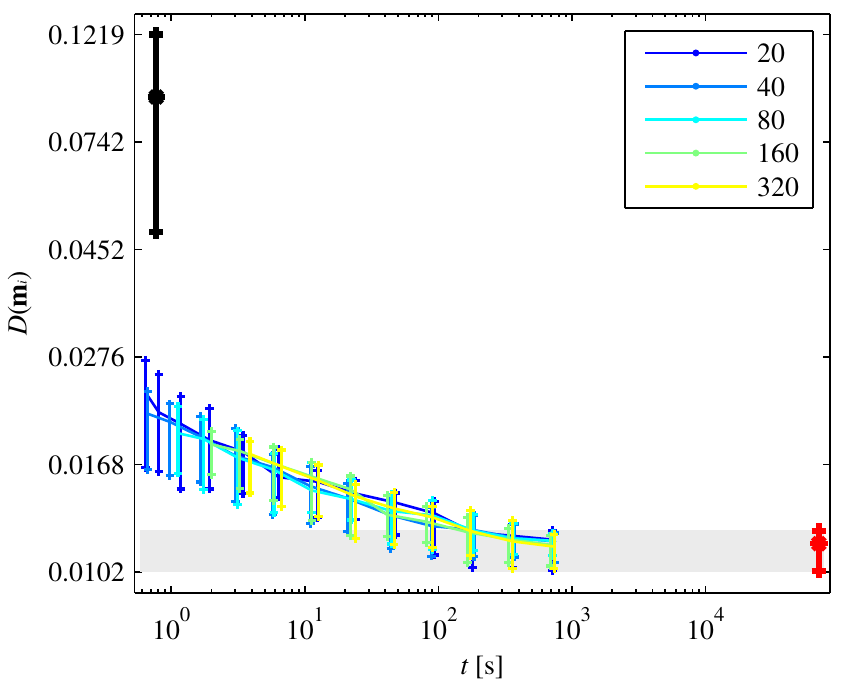}
	\caption{Effect of $\kappa$ on the \textsc{Wind} time series. Equivalent results were observed with the other time series.}
	\label{fig:resnparticles}
\end{figure}

\begin{figure}[t]
	\centering
	\includegraphics{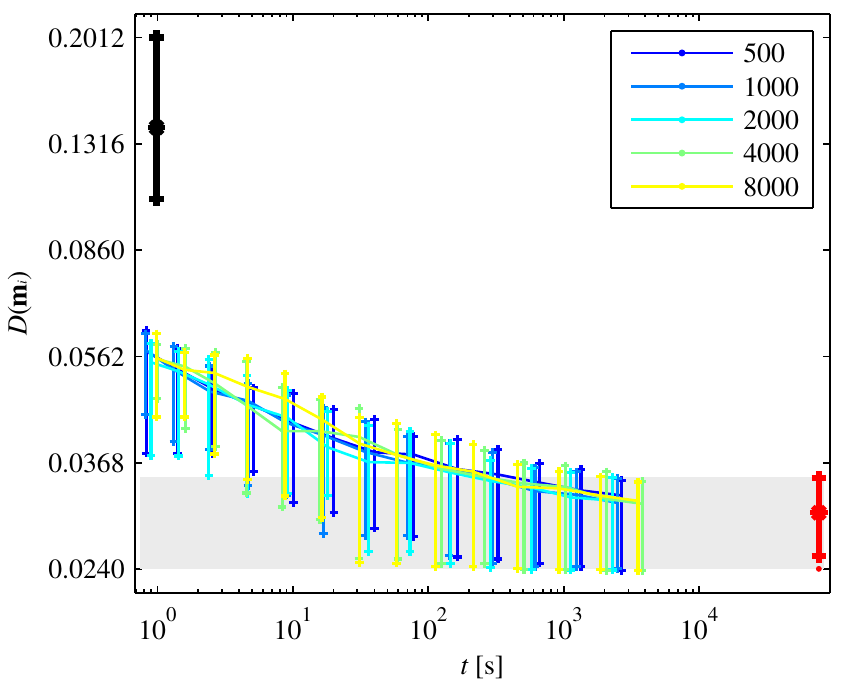}
	\caption{Effect of $\tau$ on the \textsc{FieldRecording} time series. Equivalent results were observed with the other time series.}
	\label{fig:resiterconv}
\end{figure}

Next, we studied the effect of the number of particles $\kappa$ and the stagnation threshold $\tau$. To do so, we kept the previous configuration with the local best topology and subsequently evaluated $\kappa=\{20,40,80,160,320\}$ and $\tau=\{500,1000,2000,4000,8000\}$. Essentially, we observed almost no performance changes under these alternative settings (Figs.~\ref{fig:resnparticles} and~\ref{fig:resiterconv}). We only found a statistically significant difference in the case of the \textsc{CarCount} data set. Specifically, the performance with $\tau=500$ was found to be statistically significantly worse than $\tau\geq 2000$. Regarding $\kappa$, and after considering different $n$, $w_\Delta$ and $k$, a partial tendency seemed to emerge: an increasing number of particles $\kappa$ was slightly beneficial for increasing lengths $n$, increasing $w_\Delta$, and increasing $k$. Unfortunately, we could not obtain strong empirical evidence nor formal proof for this statement. Nonetheless, in subsequent experiments, we decided to use a value for $\kappa$ and $\tau$ that dynamically adapts SWARMMOTIF's configuration to such predefined task parameters. We arbitrarily set $\tau$ proportional to $\kappa$, and $\kappa$ proportional to $n$ and in direct relation to $w_\Delta$ and $k$ (we refer to the provided code for the exact formulation). 

\begin{figure}[t]
	\centering
	\includegraphics{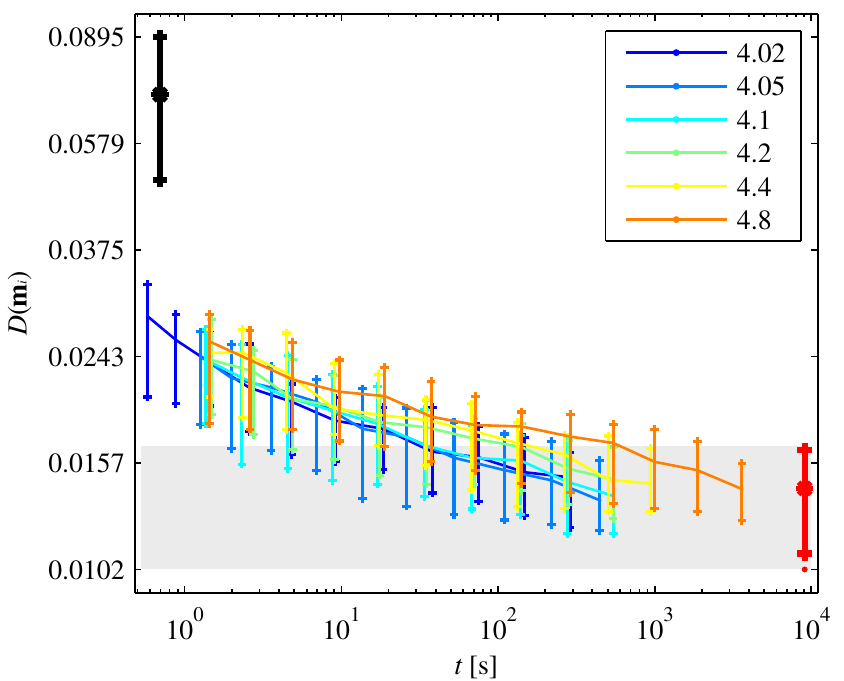}
	\caption{Effect of $\phi$ on the \textsc{Insect} time series. Equivalent results were observed with the other time series.}
	\label{fig:resphi}
\end{figure}

To conclude our pre-analysis, we studied the influence of the constriction constant $\phi$. Following common practice, we considered $\phi=\{4.02,4.05,4.1,4.2,4.4,4.8\}$. In this case, we saw that high values had a negative impact on performance (Fig.~\ref{fig:resphi}). In particular, values of $\phi\geq 4.2$ or $\phi\geq 4.4$, depending on the data set, statistically significantly increased the motif dissimilarities at a given $t$. Contrastingly, values $4<\phi<4.2$ yielded stable dissimilarities with no statistically significant variation (in some data sets, this range could be extended to $4<\phi\leq4.4$). It is well-known that higher $\phi$ values favor exploitation rather than exploration~\cite{Clerc02TEVC}. Hence, it is not strange to observe that low $\phi$ values are more appropriate for searching the large motif spaces we consider here. We finally chose $\phi=4.05$.

Overall, the result of our pre-analysis suggests a high degree of robustness with respect to the possible configurations. The topology $\theta$, the number of particles $\kappa$, the stagnation threshold $\tau$, and the constriction constant $\phi$ have, in general, no significant influence on the obtained results. The only consistent exception is observed for values of $\phi\geq 4.4$, which are not the most common practice~\cite{Parsopoulos10BOOK}. The global best and wheel topologies could also constitute a further exception. However, as we have shown, these become qualitatively equivalent to the rest as execution time $t$ increases, yielding no statistically significant difference. We believe that the reported stability of SWARMMOTIF against the tested configurations and data sets justifies the use of our setting for finding motifs in diverse time series coming from further application domains.

\subsection{Variants}
\label{sec:resvariants}

Using the configuration resulting from the previous section, we subsequently assessed the performance of the variations considered in Sec.~\ref{sec:swarmmotifvariants}. We started with the sociability variant, experimenting with social-only models, $\alpha=1$, cognitive-only models, $\alpha=0$, and intermediate configurations, $\alpha=\{0.2,0.33,0.66,0.8\}$. Apart from the fact that no clear tendency could be observed, none of the previous settings was able to consistently reach the performance achieved by the original variant ($\alpha=1/2$, Eq.~\ref{eq:c1c2}) in all time series. That is, none of the previous settings could statistically significantly outperform $\alpha=1/2$ in the majority of the data sets.

Next, we experimented with the stochastic and the velocity clamping variants. While the former did not improve our results, the latter led to a statistically significant improvement for some time series. Because of that, we decided to discard the use of a stochastic variant but to incorporate velocity clamping to our main algorithm. The former could be difficult to justify while the latter has empirical evidence behind it (Sec.~\ref{sec:swarmmotifvariants}).

\begin{figure}[t]
	\centering
	\includegraphics{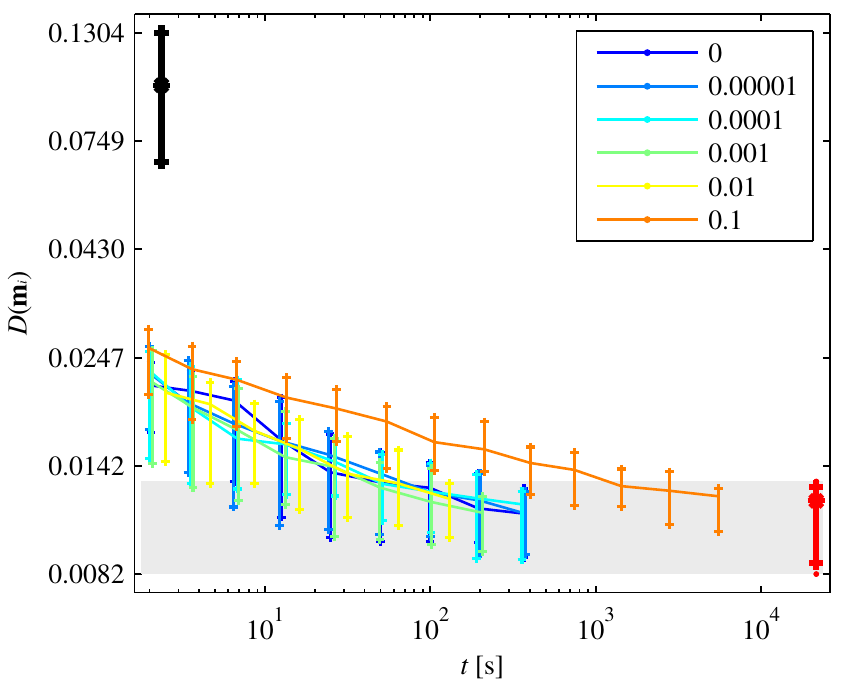}
	\caption{Effect of $\rho$ on the \textsc{EEG} time series. Equivalent results were observed with the other time series.}
	\label{fig:rescrazy}
\end{figure}

Finally, we experimented with craziness and its probability $\rho$. The results showed a similar performance for $0\leq\rho<0.001$, a slightly better performance for $0.001\leq\rho\leq 0.01$, and an increasingly worse performance for $\rho>0.01$ (Fig.~\ref{fig:rescrazy}). A statistically significant difference was found between $\rho\leq 0.01$ and $\rho> 0.1$, being $\rho> 0.1$ a consistently worse setting. These results were expected, as the swarm performs a more random search with increasing $\rho$, being completely random in the limiting case of $\rho=1$. The slightly better performance for $0.001\leq\rho\leq 0.01$ was not found to be statistically significant under our criteria. However, it was visually noticeable for some data sets. For instance, with the \textsc{EEG} data set, we see that curves 33 and 34 hit the dissimilarities of the true exact motif set $\mathcal{M}^\ast$ (gray area) two or three times earlier than curves 30, 31, and 32 (Fig.~\ref{fig:rescrazy}). With these results, and seeing that $\rho$ values between 0.001 and 0.01 never harmed the performance of the algorithm, we chose $\rho=0.002$.

\subsection{Final Performance}
\label{sec:resfinal}

\begin{figure*}[t]
	\centering
	\includegraphics{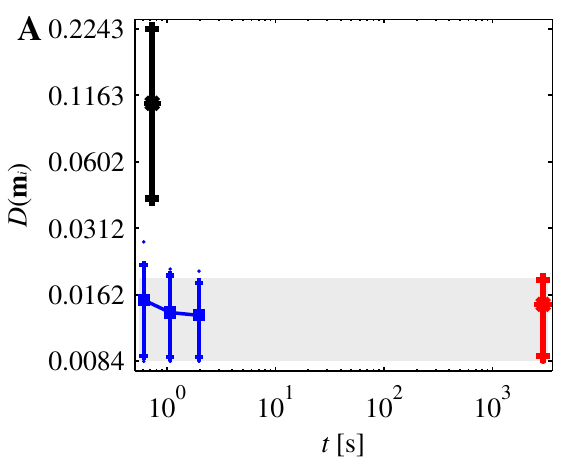} \,\,
	\includegraphics{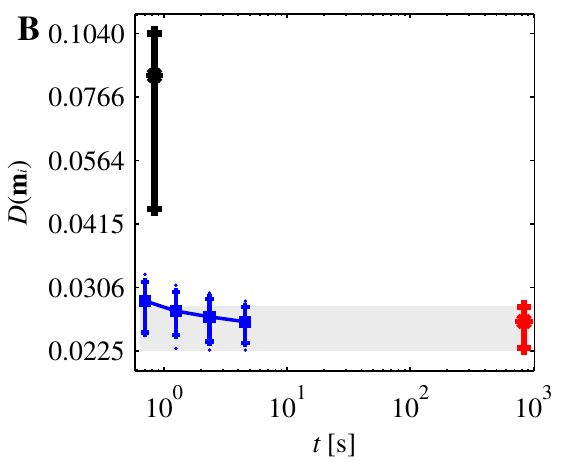} \,\,
	\includegraphics{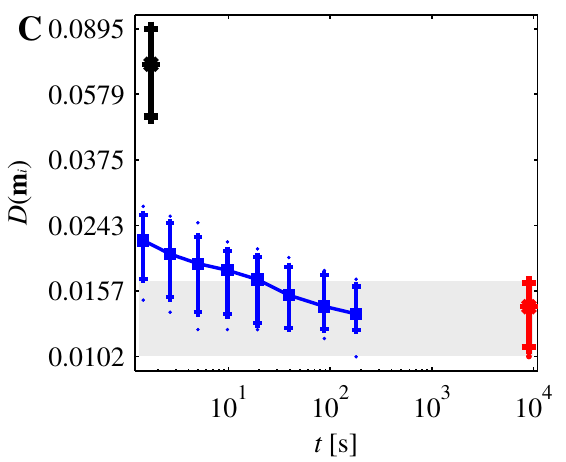} \\ \vspace*{0.3cm}
	\includegraphics{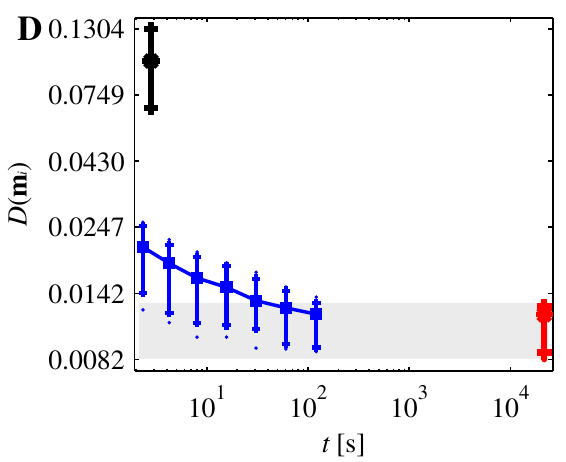} \,\,
	\includegraphics{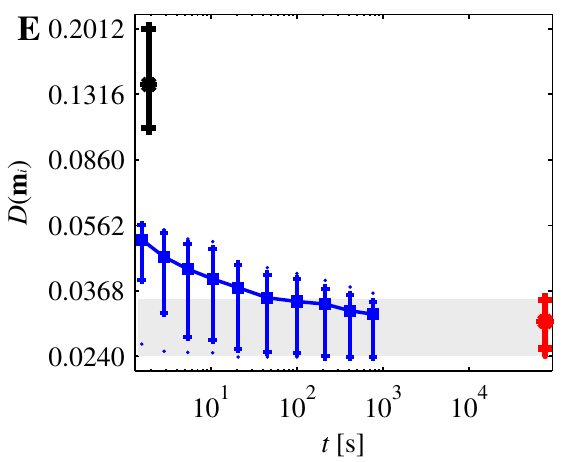} \,\,
	\includegraphics{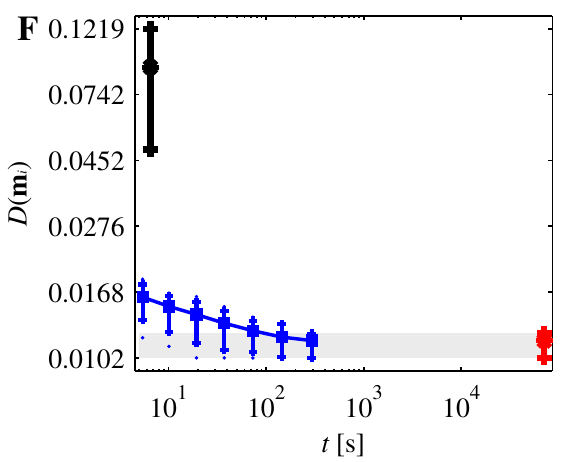} \\ \vspace*{0.3cm}
	\includegraphics{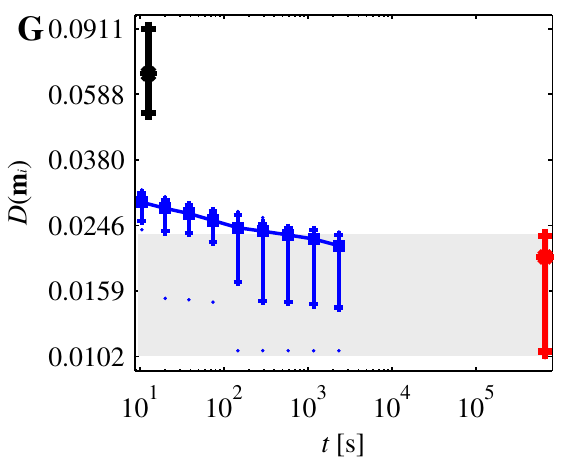} \,\,
	\includegraphics{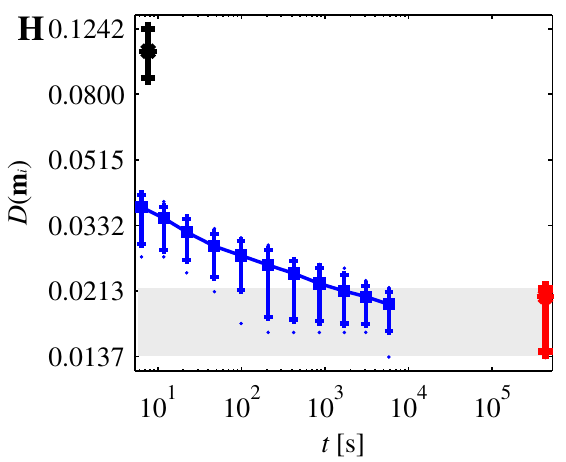} \,\,
	\includegraphics{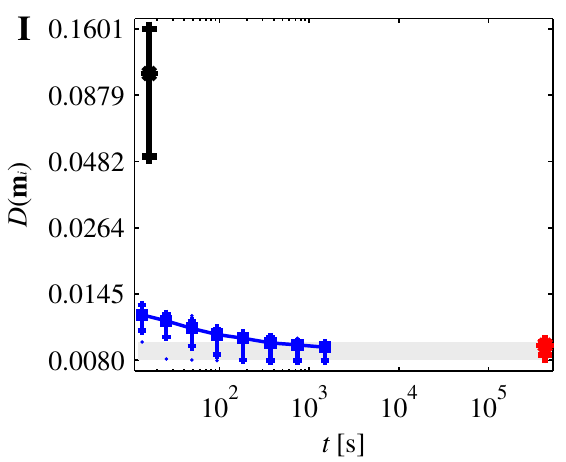}
	\caption{SWARMMOTIF performance on the considered time series: (A) \textsc{DowJones}, (B) \textsc{CarCount}, (C) \textsc{Insect}, (D) \textsc{EEG}, (E), \textsc{FieldRecording}, (F) \textsc{Wind}, (G) \textsc{Power}, (H) \textsc{EOG}, and (I) \textsc{RandomWalk}.}
	\label{fig:resfinal}
\end{figure*}

After extending SWARMMOTIF with velocity clamping and craziness, we assess its performance on all considered time series using the default parameter combination resulting from the previous two sections. As can be seen, the obtained motif dissimilarities are far from the random sampling in all cases (Fig.~\ref{fig:resfinal}; notice the logarithmic axes). In addition, we see that SWARMMOTIF is able to already obtain dissimilarities within $\mathcal{M}^\ast$ as soon as its execution begins. Specifically, motif dissimilarities in $\mathcal{M}$ start to overlap the ones in $\mathcal{M}^\ast$ at $t<10$\,s for practically all time series. The only exceptions are \textsc{EOG} and \textsc{RandomWalk}, where $\mathcal{M}$ starts to overlap with $\mathcal{M}^\ast$ at $t<100$\,s. We hypothesize that taking a smaller number of particles $\kappa$ could make $\mathcal{M}$ overlap with $\mathcal{M}^\ast$ earlier, but leave the formal assessment of this hypothesis for future work.

Finally, as execution time $t$ progresses, we see that SWARMMOTIF consistently retrieves lower dissimilarities, up to the point that $\mathcal{M}\simeq\mathcal{M}^\ast$ (Fig.~\ref{fig:resfinal} and Table~\ref{tab:result}). Following the condition we specify in Sec.~\ref{sec:eval}, this means that the distances in the motif set obtained by SWARMMOTIF are not statistically worse than the ones of the true exact motif set. With respect to MOEN's execution time, this happens 1487 (\textsc{DowJones}), 184 (\textsc{CarCount}), 50 (\textsc{Insect}), 179 (\textsc{EEG}), 100 (\textsc{FieldRecording}), 241 (\textsc{Wind}), 286 (\textsc{Power}), 74 (\textsc{EOG}), and 287 (\textsc{RandomWalk}) times faster. This implies between one and three orders of magnitude speedups (more than that for \textsc{DowJones}). Overall, we believe this is an extremely competitive performance for an anytime motif discovery algorithm.

\begin{table*}
	\centering
	\caption{Comparison between MOEN and SWARMMOTIF results. The latter are taken at the time when the defined stopping criterion is met (Sec.~\ref{sec:eval}). Results for other time steps are available online (Sec.~\ref{sec:intro}).}
	\label{tab:result}
	\resizebox{1\linewidth}{!}{
	\begin{tabular}{ll|ccccccccc}
	\hline\hline
	Approach & Result & \textsc{DowJones} & \textsc{Dodgers} & \textsc{Insect} & \textsc{EEG} & \textsc{FieldRecording} & \textsc{Wind} & \textsc{Power} 	& \textsc{EOG} & \textsc{RandomWalk} \\
	\hline
	MOEN & $t$ [s] & 2933.9 & 838.0 & 9051.1 & 21596.0 & 76415.0 & 70727.2 & 672161.9 & 430790.3 & 430676.6 \\
		& $D(\textbf{m}_i)^\ast$ median & 0.0147 & 0.0260 & 0.0140 & 0.0119 & 0.0301 & 0.0117 & 0.0199 & 0.0206 & 0.0091 \\
		& $D(\textbf{m}_i)^\ast$ min & 0.0084 & 0.0225 & 0.0102 & 0.0081 & 0.0240 & 0.0102 & 0.0102 & 0.0137 & 0.0080 \\
		& $D(\textbf{m}_i)^\ast$ max & 0.0191 & 0.0279 & 0.0168 & 0.0131 & 0.0347 & 0.0124 & 0.0232 & 0.0218 & 0.0094 \\
	\hline
	SWARMMOTIF & $t$ [s] & 2.0 & 4.6 & 180.0 & 120.6 & 764.8 & 293.1 & 2346.5 & 5807.7 & 1497.3 \\
		& $D(\textbf{m}_i)$ median & 0.0132 & 0.0259 & 0.0135 & 0.0119 & 0.0315 & 0.0117 & 0.0214 & 0.0195 & 0.0090 \\
		& $D(\textbf{m}_i)$ 5\textsuperscript{th} percentile & 0.0087 & 0.0234 & 0.0121 & 0.0090 & 0.0239 & 0.0102 & 0.0142 & 0.0163 & 0.0079 \\
		& $D(\textbf{m}_i)$ 95\textsuperscript{th} percentile & 0.0182 & 0.0278 & 0.0162 & 0.0130 & 0.0341 & 0.0122 & 0.0230 & 0.0212 & 0.0092 \\
	\hline\hline
	\end{tabular}
	}
\end{table*}

\section{Conclusion}
\label{sec:conclusion}

In this article, we propose an innovative standpoint to the task of time series motif discovery by formulating it as an anytime multimodal optimization problem. After a concise but comprehensive literature review, we reason out the new formulation and the development of an approach based on evolutionary computation. We then highlight the several advantages of this new formulation, many of which relate to a high degree of flexibility of the solutions that come from it. To the best of our knowledge, such a degree of flexibility is unseen in previous works on time series motif discovery. 

We next present SWARMMOTIF, an anytime multimodal optimization algorithm for time series motif discovery based on particle swarms. We show that SWARMMOTIF is extremely competitive when compared to the best approach we could find in the literature. It obtains motifs of comparable similarities, in considerably less time, and with minimum memory requirements. This is confirmed with 9~independent real-world time series of increasing length coming from a variety of domains. Besides, we find that the high majority of the possible implementation choices lead to non-significant performance changes in all considered time series. Thus, given this robustness, we can think about the proposed solution as being parameter-free from the user's perspective. Overall, if we add the aforementioned, unprecedented degree of flexibility, SWARMMOTIF stands out as one of the most prominent choices for motif discovery in long time series streams. Since the used data and code are available online (Sec.~\ref{sec:intro}), the research presented here is fully reproducible, and SWARMMOTIF is freely available to researchers and practitioners.

We believe that the consideration of multimodal optimization algorithms is a relevant direction for future research in the field of time series analysis and mining. Not only with regard to motif discovery, but also in other tasks such as querying for segments of unknown length~\cite{Kahveci04TKDE} or determining optimal alignments and similarities~\cite{Serra14KBS}. With regard to the latter, we envision powerful approaches to variable-length local similarity calculations, in the vein of existing dynamic programming approaches~\cite{Smith81JMB,Serra09NJP}. Finally, we believe that considering the search spaces and the time constraints derived from time series problems can be a challenge for the evolutionary computation community. We look forward to exploring all these topics in forthcoming works, together with other multimodal optimization techniques that could be easily mapped to the problem of time series motif discovery. 

\section*{Acknowledgment}

We would like to thank all the people who contributed the data sets used in this study and Abdullah Mueen for additionally sharing his code. We would also like to thank Xavier Anguera for useful discussions that motivated the present work. This research has been funded by 2009-SGR-1434 from Generalitat de Catalunya, JAEDOC069/2010 from Consejo Superior de Investigaciones Cient\'ificas (JS), TIN2012-38450-C03-03 from the Spanish Government, and E.U.\ Social and FEDER funds (JS).

\ifCLASSOPTIONcaptionsoff
  \newpage
\fi



\bibliographystyle{IEEEtran}
\bibliography{IEEEabrv,bibjserra}
%

%


\begin{IEEEbiography}[{\includegraphics[width=1in,height=1.25in,clip,keepaspectratio]{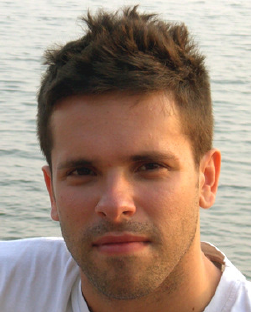}}]{Joan Serr\`a} is a postdoctoral researcher with the Dept.\ of Learning Systems of IIIA-CSIC, the Artificial Intelligence Research Institute of the Spanish National Research Council in Bellaterra, Barcelona, Spain. His research focuses on machine learning, data mining, time series, and complex networks with applications to computer audition, music information retrieval, multimedia annotation and, to a much lesser extent, the social and health sciences. He obtained his MSc (2007) and PhD (2011) in Computer Science from Universitat Pompeu Fabra, Barcelona, where he was also an adjunct professor with the Dept.\ of Information and Communication Technologies (2006--2011). He has been involved in more than 10 publicly-funded research projects and co-authored over 60 publications, some of them highly-cited and in top-tier journals and conferences of diverse scientific areas. He also regularly acts as a reviewer for some of those and other publications.
\end{IEEEbiography}


\begin{IEEEbiography}[{\includegraphics[width=1in,height=1.25in,clip,keepaspectratio]{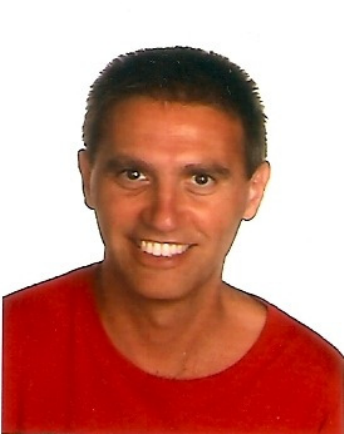}}]{Josep Lluis Arcos} is a research scientist of the Artificial Intelligence Research Institute of the Spanish National Research Council (IIIA-CSIC), where he is a member of the Learning Systems Department. He received the MSc (1991) and PhD (1997) in Computer Science from the Technical University of Catalonia, Spain. He also received a MSc in Sound and Music Technology (1996) from the Pompeu Fabra University, Spain. He is co-author of more than 125 scientific publications with contributions on machine learning, case-based reasoning, multi-agent systems, self-organizing systems, or AI and music. He is co-recipient of several awards at case-based reasoning and computer music conferences. Presently, he is working on machine learning and on its applications to health care, Internet, and music domains. He acts as a reviewer of several international journals and of top international conferences.
\end{IEEEbiography}




\vfill


\end{document}